\documentclass[11pt]{article}

\usepackage[margin=1in]{geometry}
\usepackage{graphicx}
\usepackage{booktabs}
\usepackage{array}
\usepackage{float}
\usepackage{caption}
\usepackage{microtype}
\usepackage{setspace}
\usepackage{hyperref}
\usepackage[T1]{fontenc}
\usepackage[utf8]{inputenc}
\usepackage{newtxtext,newtxmath}
\usepackage{xurl}

\hypersetup{colorlinks=true,linkcolor=black,citecolor=black,urlcolor=blue}
\title{Modeling Human Behavior in the Prisoner's Dilemma with Large Language Models}

\author{
Raina Gao$^{1}$ \qquad Tristan Raffo$^{1}$
}

\date{
\small $^{1}$Thomas Jefferson High School for Science and Technology
}

\begin{document}
\maketitle

\begin{abstract}
The Prisoner's Dilemma has long served as a foundational framework for studying human decision-making and cooperative outcomes across numerous social scenarios, yet human behavior consistently deviates from predictions of fully rational economic theory. Previous works have extensively studied human and LLM outcomes in the Prisoner's Dilemma; however, these studies have lacked at least one key feature: they either did not include a memory component for the LLM agents, did not use accessible open-source models, or did not compare models against empirical human data. This project investigates whether open-source LLMs, TII Falcon 7B Instruct, Google Gemma 7B, Meta Llama 3.1 8B Instruct, Microsoft Phi Mini MoE Instruct, and Qwen 2.5 7B Instruct, can accurately model human cooperation patterns in repeated Prisoner's Dilemma games, while also examining how an LLM's knowledge that its opponent will be shuffled affects its decisions. Using a standardized prompting framework to match established Prisoner's Dilemma experiments, we simulated repeated interactions in two trial types: a fixed-partner setting where both agents retain memory, and a shuffled-partner setting where a memory-enabled agent, informed that its opponent would change across rounds, interacts with a memoryless agent. By testing both interactions, this study evaluates how varying agent environments affect cooperation rates, and broadly, decision-making behaviors. For each trial, we measured cooperation rates over 100 rounds and 10-round increments and compared to baseline human data from Montero-Porras et al. (2022). Under both frameworks, all LLMs differed from the human cooperation rate. Open source LLMs generally over-cooperated with respect to humans, particularly in the shuffled framework, where human cooperation rates averaged 28.5\% while several models exceeded 80\%. Qwen-2.5-7B-Instruct showed evidence of deterministic tendencies, while others, like Phi-mini-MoE-Instruct, displayed much more variability, and even the ability to approximate human cooperation rates briefly. These findings suggest that LLMs have potential as behavioral simulation agents and provide a benchmark across two agent-interaction environments, although current open-source models are not yet reliable substitutes for human participants.
\end{abstract}

\section{Introduction}
The Prisoner's Dilemma has long served, in economic and behavioral studies, as a model of human decision-making and cooperative outcomes that can be applied to numerous social scenarios. Economic theory dictates how games should proceed under the assumption of perfectly rational players, yet empirical research has repeatedly shown that humans do not consistently behave as payoff-maximizing \textit{Homo economicus} agents. Human decisions are influenced by factors such as prior interactions, expectations of others, and willingness to cooperate, making observed behavior substantially more complex than the predictions of simple rational-choice models. As a result, experimental Prisoner's Dilemma data can provide useful insight into cooperation, strategic decision-making, and repeated social interactions.

Large language models (LLMs) have increasingly been studied as potential simulations of human behavior. Through training on large quantities of human-generated text, LLMs learn patterns in language, reasoning, and decision-making that may allow them to generate responses resembling those of human participants. This has created growing interest in using LLM agents in behavioral and social-science experiments, particularly because simulated participants can be generated at a scale that would be difficult to achieve with human subjects. Human experiments require participant recruitment, compensation, scheduling, and other resources, while LLM-based experiments can conduct large numbers of standardized interactions computationally.

Previous studies have already examined LLM behavior in the Prisoner's Dilemma and other economic games, suggesting that language models can exhibit cooperation, defection, and other strategic behaviors observed in human experiments. However, existing work leaves several important limitations. Studies frequently use large proprietary models, which may be costly, inaccessible for local experimentation, or difficult for researchers to reproduce. Other studies evaluating open-source models do not necessarily compare their behavior directly with empirical human data. Additionally, many evaluations treat Prisoner's Dilemma decisions as isolated interactions, despite the importance of memory and previous outcomes in repeated human decision-making. Thus, prior work generally lacks at least one of three important components: accessible open-source models, an explicit memory mechanism across interactions, or direct comparison with empirical human behavior.

This project addresses these limitations by evaluating whether locally runnable, open-source LLMs can reproduce patterns of human behavior in repeated Prisoner's Dilemma interactions. We test multiple publicly accessible models, including TII Falcon 7B Instruct, Google Gemma 7B, Meta Llama 3.1 8B Instruct, Microsoft Phi Mini MoE Instruct, and Qwen 2.5 7B Instruct. LLM agents participate in simulated Prisoner's Dilemma games in which their previous interactions can be retained as memory, allowing their decisions to respond to prior cooperation and defection rather than treating every round independently. Their resulting cooperation patterns are then compared with empirical human Prisoner's Dilemma data to evaluate how closely each model reproduces observed human behavior.

Rather than determining only whether an LLM can make a plausible individual Prisoner's Dilemma decision, this study evaluates whether its broader behavioral patterns remain human-like across repeated interactions. Comparing multiple open-source models also allows us to examine whether human-like behavior is consistent across model architectures or concentrated in particular models. The memory component further tests whether providing models with interaction history improves behavioral similarity by more closely reproducing the information available to humans during repeated games.

The primary contribution of this project is therefore an empirical evaluation of accessible open-source LLMs as potential behavioral agents in repeated economic-game research. By jointly incorporating model memory, multiple locally runnable LLMs, and comparison against real human outcomes, the study addresses limitations that have generally been examined separately in previous work. If smaller open-source models can reproduce meaningful patterns of human cooperation, they could provide researchers with a more accessible and reproducible method for conducting large-scale simulations of strategic behavior. At the same time, differences between LLM and human behavior can identify important limitations in treating language models as substitutes for human participants, establishing where they may be useful as behavioral simulations and where direct human experimentation remains necessary.

\section{Related Works}
Human behavior has been studied extensively in the Prisoner's Dilemma, despite challenges in coordinating subjects across time zones, geographic locations, and availability. Montero-Porras et al. (2022) conducted a long-term study over one hundred rounds with human participants, both in shuffled and fixed-partner frameworks, analyzing patterns in cooperation and strategy, noting that memory was an important factor in forming strategies over time. Grujic et al. similarly studied 169 volunteers in both shuffled and fixed settings in 2010, but an important drawback of human studies, as demonstrated by their work, was the requirement of an incentive, monetary or otherwise, to attract human participants. Thus, there opens a gap for LLMs to be used in behavioral research rather than humans, if they are able to mimic human behavior accurately -- LLMs require no incentive to participate. Indeed, Gao et al. (2023) highlight how LLMs are growing in application to research with economic and behavioral games, but there is much progress to be made on the side of mimicking human behavior: relatively few studies have incorporated a memory component into the LLMs used; even further, the nature of LLMs is that they are familiar with economic theory and strategy to an extent beyond that of the average layperson, making them unrepresentative of humans; lastly, as shown by Fontana et al. (2024), some frontier LLMs cooperate more frequently than humans do. It is known, as mentioned in Camerer (2011), that humans deviate from theoretical equilibria and of course, rationality: LLMs used for behavioral research thus ought to replicate irrational behavior while staying true to human skepticism. LLMs used for behavioral research also ought to be fairly accessible, especially to keep costs lower than the sum of incentives and planning involved with human subjects: for this reason, despite many studies around frontier models, we decide to test smaller, open-source models.

Chen et al. (2023) confirm that LLMs emerge as more economically rational than humans, which is a benefit for aiding human decision making, but means that LLMs used to reduce difficulties in simulating human behavioral research could require directly induced irrationality. Instead, avoiding this technique, Chen et al. (2023) illustrate an alternative approach to reduce the rationality of LLMs: alternative problem/question framing. The notion is that concealing the situation to the LLM, by representing payoff and choices alternatively with respect to the traditional game theoretic units and options, reduces the chances of the LLM recognizing it is in a game theoretic scenario and thus acts more irrationally, just as many humans do not expect or recognize an application of game theory when they are not told they are in a Prisoner's Dilemma. A need for alternative framing makes the experimental setup from Montero-Porras et al. (2022) appealing in benchmarking the potential for LLMs to mimic irrational behavior seen in humans: rather than representing the problem as a Prisoner's Dilemma, the LLM may be given the choice of yellow or blue, with no hinting towards game theory. Introducing a memory component, which can be done through the prompt side, addresses some gaps from Gao et al. (2023); introducing an alternative framing, also done via the prompt side, leverages the ability to ``induce'' irrationality as shown by Chen et al. (2023). A study with both of these components would be taking a step towards mimicking human behavior, as we seek to do. Lu et al. (2025) reiterate that prompt-side adjustments to mimic human behavior are indeed valid and effective, meaning that the desired effects may indeed be obtained without modifying model parameters or internal workings.

\section{Methods}
\subsection{Overview}
We evaluated whether open-source large language models (LLMs) reproduce human cooperation patterns in repeated Prisoner's Dilemma interactions. Using TII Falcon, Google Gemma, Meta Llama, Microsoft Phi, and Qwen families via Hugging Face Transformers library, each model received a standardized prompt including description of its role, a payoff matrix for each choice, and an objective of maximizing total points across 100 rounds. To reduce framing effects, each LLM was presented with available actions as the colors blue and yellow rather than revealing the game as the Prisoner's Dilemma. We conducted two trial types: a fixed-opponent condition and a shuffled-opponent condition.

\begin{table}[H]
\centering
\caption{Prompt for the fixed-opponent condition. The shuffled-opponent condition instead informed the memory-enabled agent that its opponent would change across rounds and did not provide opponent-specific interaction history.}
\label{tab:prompt}
\begin{minipage}{0.94\textwidth}
\small
\ttfamily
You are playing a repeated two-option game for points across many rounds. In each round, you and the SAME other player choose simultaneously.\\[0.4em]
Actions: B = Blue; Y = Yellow\\[0.4em]
Points to YOU:\\
- If you choose B and the other chooses B: 3\\
- If you choose B and the other chooses Y: 0\\
- If you choose Y and the other chooses B: 4\\
- If you choose Y and the other chooses Y: 1\\[0.4em]
However, you are a human decision-maker with limited ability to compute optimal strategies. You do not assume the other player follows a fixed rule, and you may use a mixed strategy. History (most recent first, for reference only, not a rule to follow):\{history\_block\}\\[0.4em]
Based on your overall impression of the interaction so far, choose your action for this round. Choose your action for this round ONLY to maximize your total points over all rounds Respond with exactly ONE character: B or Y. Do not include any other text.
\end{minipage}
\end{table}

\subsection{Prompting Structure}
We created a Prisoner's Dilemma with the actions C (cooperate) and D (defect). To imitate Montero-Porras et al. (2022), we adopted the same payoff matrix in terms of ``points'' as follows:

\begin{table}[H]
\centering
\caption{Payoff Matrix}
\label{tab:payoff}
\begin{tabular}{c|cc}
\toprule
(Player 1, Player 2) & C & D \\
\midrule
C & (3, 3) & (0, 4) \\
D & (4, 0) & (1, 1) \\
\bottomrule
\end{tabular}
\end{table}

where the conventional Nash Equilibrium would dictate a dominant strategy for both players to defect, leading to an equilibrium payoff of one point per player per round. We, however, do not make the points - currency conversion they made due to LLMs needing no incentive to play the game, another benefit of this method. The game, when presented to the LLM, is portrayed just as Montero-Porras et al. (2022) and Grujić et al. (2010) portrayed it -- the choice C was represented by the color blue, and the choice D by the color yellow. The game was never referred to as ``Prisoner's Dilemma'' when given to the LLM, and the scenario of a prison or crime conviction is not mentioned either, to prevent unwanted bias due to recognizing the scenario. The LLM was given the task of maximizing its payoff over the one hundred rounds it played by choosing whichever strategy it sees fit.

Through importing Hugging Face's Transformers library, we utilized a pipeline to prompt each selected LLM. Each model-condition trial consisted of one hundred rounds; the prompt included a thorough description of the game's scenario (making a choice against another agent), and the payoffs for each combination of choices. The LLM was prompted to limit its response to a one-token, binary B or Y representing blue or yellow. When the response strayed away from an exact B or Y, we repeated the prompt until it yielded a B or Y and discarded any invalid responses. We tracked the responses of both agents in each round and paired them to determine the resulting payoff.

% TODO BEFORE SUBMISSION: Add the exact generation settings used (temperature, do_sample, seed, top_p/top_k if applicable) and the number/frequency of invalid-output retries. These values are not available in the current manuscript and should not be guessed.

With both choices in the round decided for each model, we pair the choices to determine payoff. Payoffs were determined by comparing the two choices (Blue + Yellow, BB, YB, YY). The choice and payoff for each agent are stored in the output file; in the fixed-partner scenario, the prompt contains knowledge about previous decisions from both the agent and its opponent. In the shuffled-partner scenario, a memory-enabled agent is informed that its opponent changes across rounds and interacts with a memoryless agent, preventing reliance on a persistent opponent-specific history. This served as a human-like ``memory'' aspect for the LLM, which was able to reference previous results in the fixed-partner condition. The ``memory'' included each previous scenario with the format: ``In round [n-1...n-2...n-3...1], you chose [choice], opponent chose [opponent choice], payoff: [calculated payoff],'' where n represents the current round number.

\subsection{Outcome Measures}
Model cooperation rates were compared with the human results reported by Montero-Porras et al. (2022). We examined both aggregate cooperation and changes in cooperation across the 100 rounds. Ten-round rolling averages were used to display longer-term patterns while reducing round-to-round variation in the figures. The data did not meet the assumptions required for the statistical tests considered.

We evaluated qualitative behavioral patterns across models and conditions, including near-deterministic strategies, sustained high or low cooperation, changes in behavior over time, and the extent to which each model approached or diverged from the human trajectory. The complete round-level outputs are retained in CSV format for the remaining analyses and figures.

\section{Results}
\subsection{Overall Cooperation Rates}
Across both experimental conditions, the LLMs generally exhibited higher cooperation rates than the human comparison group. This difference was especially pronounced in the shuffled-opponent condition, where most models maintained high levels of cooperation across the 100 rounds. In the fixed-opponent condition, aggregate LLM cooperation was slightly lower, while the human cooperation rate followed the opposite pattern and increased relative to the shuffled condition.

\begin{figure}[H]
\centering
\includegraphics[width=0.98\textwidth]{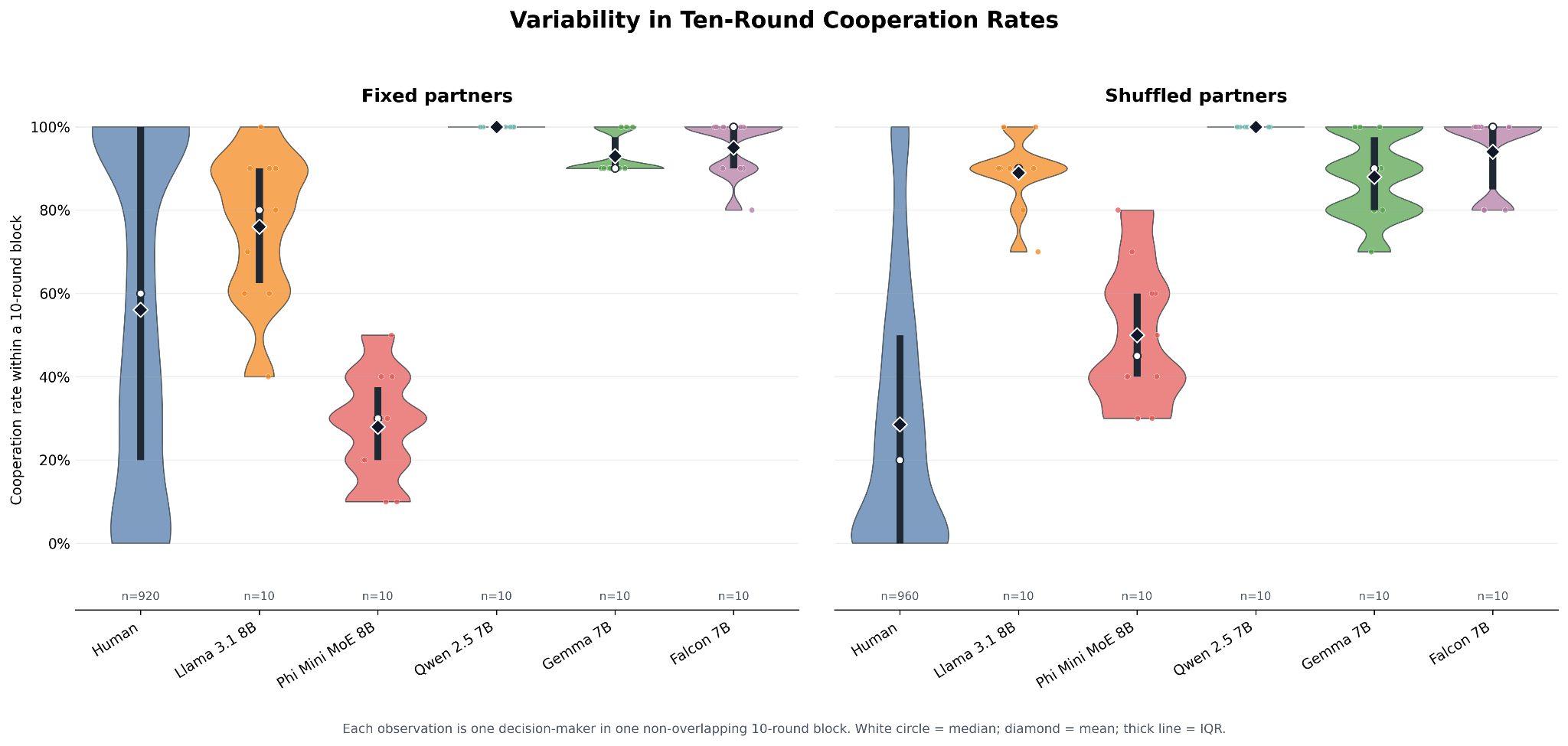}
\caption{Variability in Ten-Round Cooperation Rates. Violin width represents the frequency of observed cooperation rates, while markers show the mean, median, and interquartile range. Humans, Phi, and Llama exhibited greater variability in the fixed setting than Qwen, Gemma, and Falcon, with humans being the most variable; in the shuffled setting, humans and Phi again were the more variable groups, and the other four were fairly similar apart from the deterministic Qwen.}
\label{fig:variability}
\end{figure}

Human cooperation varied widely across participants and time in both conditions. In contrast, Qwen showed no meaningful variation because it cooperated at a rate of 100\% throughout the experiment. Gemma and Falcon also displayed limited variability and remained concentrated near very high cooperation levels. Llama showed moderate variation, still being highly cooperative, while Phi was the most variable model. Under shuffled partners, Phi's cooperation distribution shifted upward, suggesting that its behavior changed more strongly across conditions than that of the other models.

\subsection{Cooperation over time}
\begin{figure}[H]
\centering
\includegraphics[width=0.95\textwidth]{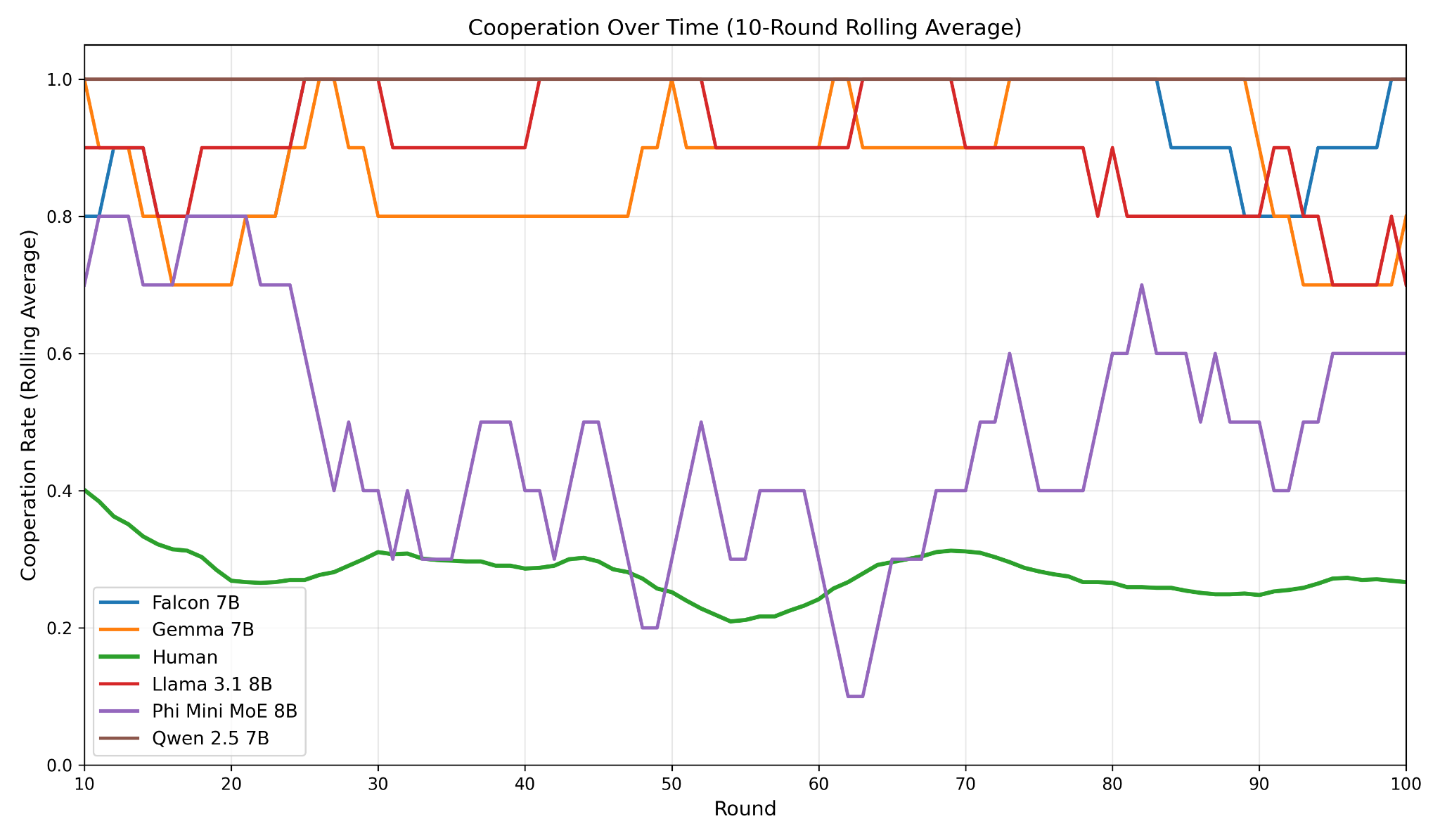}
\caption{Cooperation Over Time with Shuffled Opponents and Human Data}
\label{fig:shuffled}
\end{figure}

Figure~\ref{fig:shuffled} presents the 10-round rolling average cooperation rates in the shuffled-opponent condition. Falcon, Gemma, and Qwen maintained consistently high cooperation rates and frequently approached fully cooperative behavior. Llama also remained substantially above the human comparison rate for most rounds, although its cooperation rate varied more over time. Phi displayed the lowest cooperation rates among the tested models and showed greater fluctuation across the trial, but also approached the human cooperation rates around the middle of the trial.

\begin{figure}[H]
\centering
\includegraphics[width=0.95\textwidth]{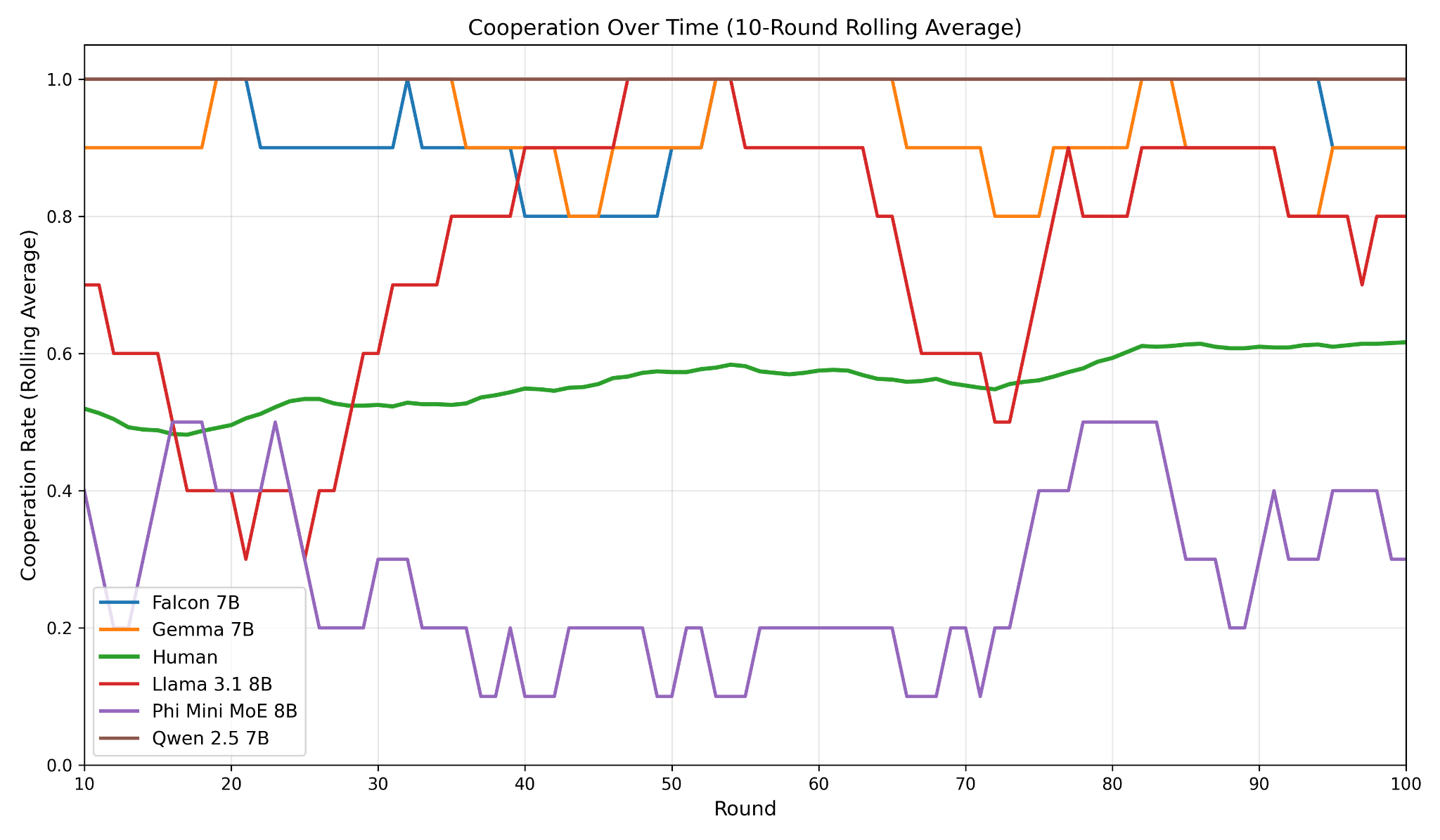}
\caption{Cooperation Over Time with Fixed Opponents and Human Data}
\label{fig:fixed}
\end{figure}

Figure~\ref{fig:fixed} presents the corresponding rolling averages for the fixed-opponent condition. Falcon, Gemma, and Qwen again approached near-deterministic strategies. However, aggregate LLM cooperation was somewhat lower than in the shuffled-opponent condition. In contrast, the human cooperation rate was higher under the fixed-opponent setting. Llama ended the trial near the human cooperation rate, but remained considerably more cooperative than humans during most earlier rounds; Phi did not come close to the human cooperation rate aside from in rounds 10-20.

\subsection{Cooperation by Model}
\begin{figure}[H]
\centering
\includegraphics[width=0.82\textwidth]{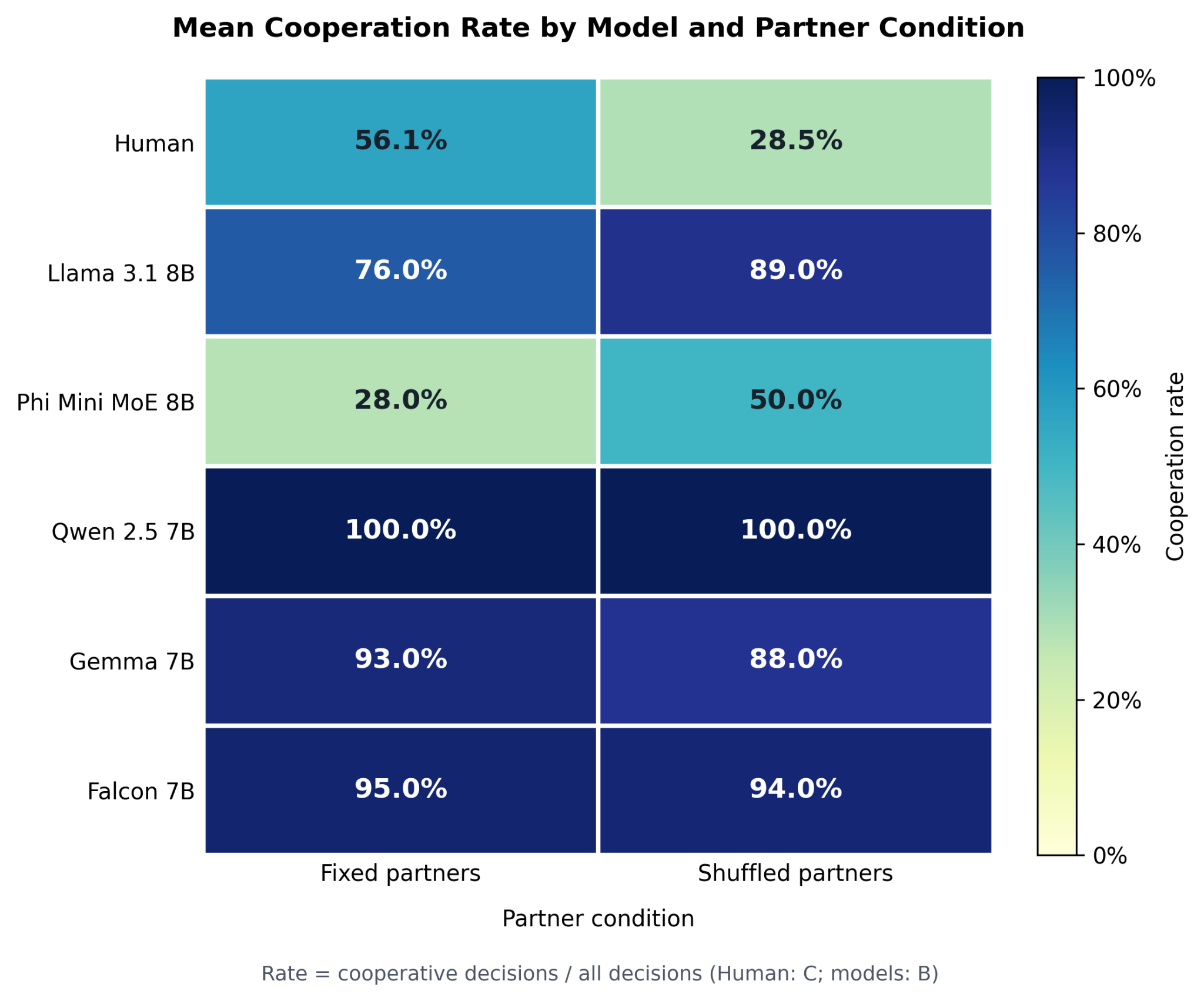}
\caption{Mean Cooperation Rate by Model and Partner Condition}
\label{fig:heatmap}
\end{figure}

Cell values show the percentage of cooperative decisions made by humans and each LLM across all rounds. Most models remained more cooperative than humans in both conditions. Humans cooperated more with fixed partners than shuffled partners, with the mean rate falling from 56.1\% to 28.5\%. Most LLMs showed the opposite pattern or remained highly cooperative in both settings. Qwen cooperated in every round under both conditions, while Gemma and Falcon also maintained rates above 87\%. Llama cooperated more with shuffled partners than fixed partners, and Phi showed the largest condition-based increase, rising from 28.0\% to 50.0\%. Overall, the models were substantially more cooperative than humans, especially when partners were shuffled. The largest decrease in cooperation when moving from fixed partners to shuffled partners was in Gemma's cooperation rates, with a 5\% decrease to 88\% from 93\%.

\subsection{Payoff Comparison}
\begin{figure}[H]
\centering
\includegraphics[width=0.95\textwidth]{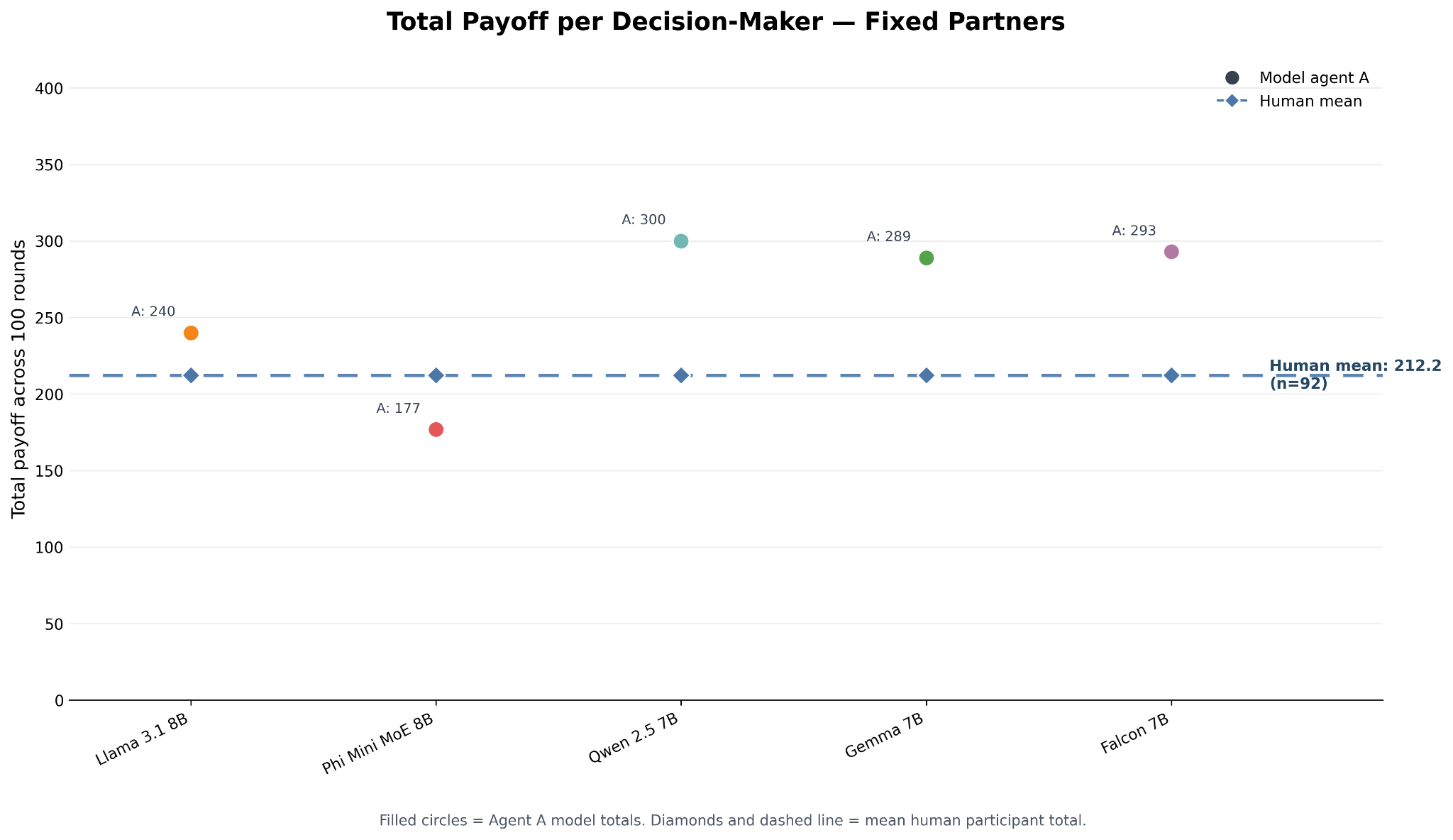}
\caption{Mean total human payoff versus model agent total payoff in fixed partners}
\label{fig:fixedpayoff}
\end{figure}

Points show individual agent payoffs for each LLM in the 100 rounds of the fixed partner setting. Qwen, cooperating every time, achieved a cumulative payoff of 300 units. Falcon and Gemma, also stably cooperative, achieved total payoffs of 293 and 289 units respectively, following close behind. Llama netted 240 units of total payoff; Phi achieved the least cumulative payoff (177 units). The mean cumulative payoff for each human player in the fixed setting in Montero-Porras et al. (2022) was 212.2 units with n = 92 subjects.

\begin{figure}[H]
\centering
\includegraphics[width=0.95\textwidth]{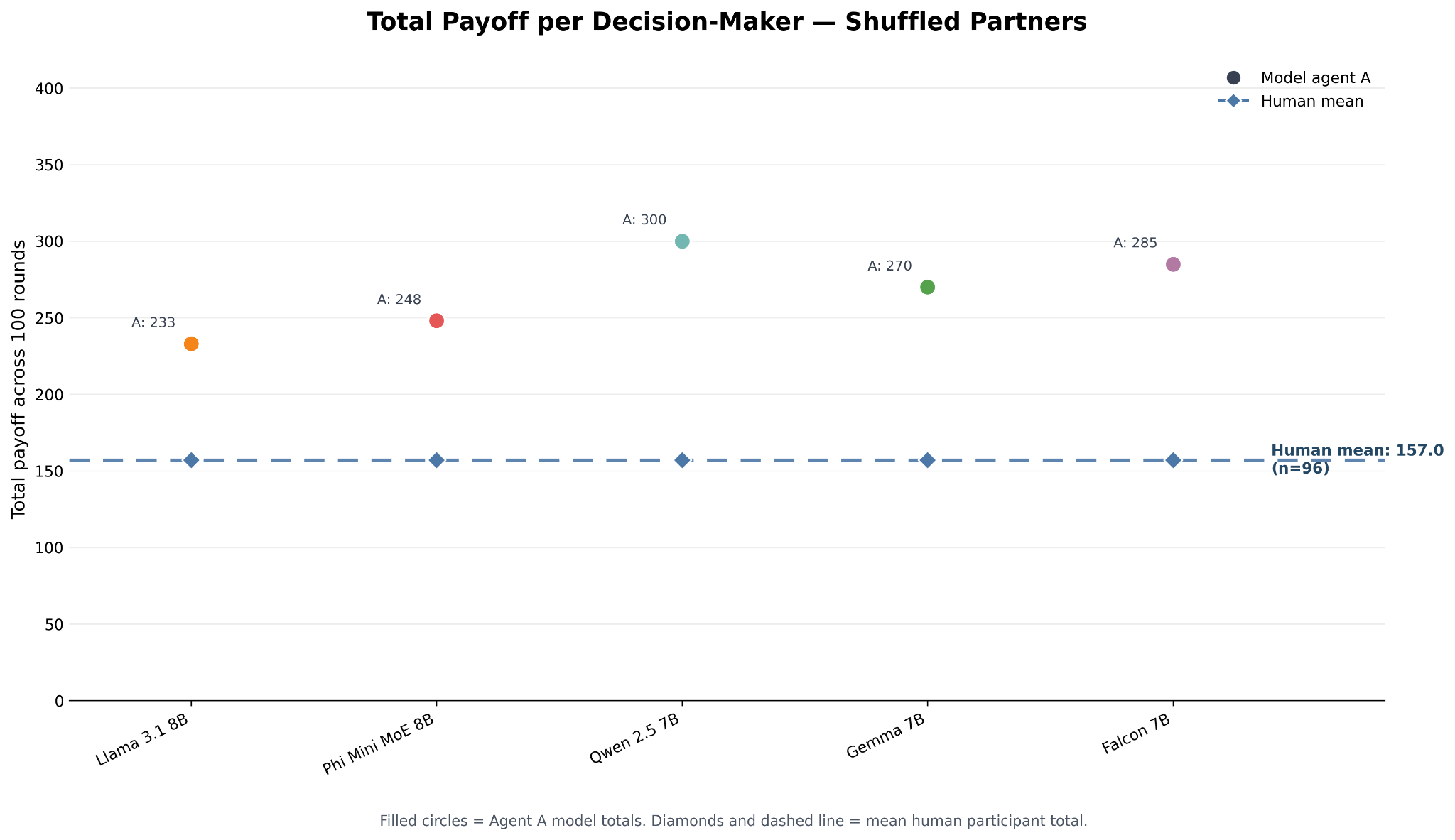}
\caption{Mean total human payoff versus model agent total payoff in shuffled partners}
\label{fig:shuffledpayoff}
\end{figure}

Qwen once again achieved a payoff of 300 total units through perfect cooperation. Falcon this time followed closely behind, with a cumulative payoff of 285 units, while Gemma obtained 270 total units of payoff. Phi did not have the lowest payoff this time -- its 248 units beat out Llama's 233 units. The human round had four more participants in the shuffled opponent setting than in the fixed setting, so the mean with n = 96 was 157 units of payoff, less than all five LLM agents.

\section{Discussion}
The results indicate that the tested open-source LLMs did not reproduce the cooperation patterns observed in human participants, particularly in how behavior changed between fixed- and shuffled-partner environments. Human cooperation decreased substantially from 56.1\% with fixed partners to 28.5\% with shuffled partners, consistent with the idea that repeated interaction with the same opponent can support reciprocal cooperation. Most LLMs did not reproduce this response. Falcon, Gemma, and Qwen remained highly cooperative regardless of partner condition, while Llama and Phi actually cooperated more when partners were shuffled. This suggests that simply giving an LLM information about the interaction environment is not sufficient to reproduce the behavioral effect that the environment has on humans. Instead, several models appear to possess strong default tendencies toward cooperation that persist even when the strategic context changes.

The variation among models is also important. Qwen exhibited completely cooperative behavior in both conditions, while Falcon and Gemma were similarly concentrated near very high cooperation rates. In contrast, Phi and Llama demonstrated substantially more variation across rounds, with portions of their trajectories approaching the human cooperation rate. Phi was particularly notable because its fixed-partner aggregate cooperation rate was relatively close to the human comparison, although its behavior diverged substantially in the shuffled condition. These results suggest that evaluating only an aggregate cooperation rate can obscure important differences in behavioral dynamics. A model may briefly approximate human cooperation while still failing to reproduce how human behavior evolves across repeated interactions or responds to changes in opponent structure.

The persistent over-cooperation observed in several models is consistent with previous work finding that LLMs may behave more cooperatively than humans in the Prisoner's Dilemma. Importantly, the alternative blue/yellow framing and prompt-based instructions intended to reduce explicit game-theoretic reasoning did not eliminate this difference. This suggests that the discrepancy cannot necessarily be resolved simply by concealing the formal name of the game or instructing the model to behave as a human decision-maker. The findings therefore place an important limitation on the use of current open-source LLMs as direct substitutes for human experimental participants: producing plausible individual decisions is not equivalent to reproducing the distribution, variability, and context-sensitive behavior observed in humans.

These conclusions should nevertheless be interpreted within the scope of the experiment. Cooperation was evaluated against one established human dataset and one payoff structure, and model behavior may be sensitive to prompting, generation parameters, model architecture, and repeated sampling. Additionally, the models do not experience monetary incentives or social consequences comparable to those faced by human participants. Future work could test multiple prompt formulations, replicate each model-condition experiment across independent runs, examine additional open-source and frontier models, and evaluate behavioral similarity using measures beyond mean cooperation. More extensive comparisons of conditional responses---such as how likely an agent is to cooperate following an opponent's cooperation or defection---could also determine whether models reproduce human strategies even when their overall cooperation rates differ.

\section{Conclusions}
This study evaluated whether accessible open-source LLMs can reproduce human cooperation behavior in a repeated Prisoner's Dilemma under fixed- and shuffled-partner conditions. Across the tested models, cooperation generally differed substantially from the human comparison data. Humans cooperated considerably more with fixed partners than with shuffled partners, whereas most LLMs either remained highly cooperative across both environments or increased cooperation when partners were shuffled. Falcon, Gemma, and especially Qwen displayed highly stable cooperative behavior, while Phi and Llama showed greater variability and occasionally approached the human trajectory. Thus, some models reproduced isolated features of human behavior, but none consistently captured both human cooperation levels and the behavioral response to changing partner conditions.

The findings provide a benchmark for the use of smaller open-source LLMs as agents in repeated behavioral experiments while demonstrating an important limitation of treating them as replacements for human participants. Incorporating interaction history, alternative game framing, and different opponent environments was not sufficient to make the tested models consistently human-like. Rather than showing that open-source LLMs can presently substitute for humans, the results suggest that behavioral fidelity must be evaluated directly and at multiple levels---including aggregate choices, variability over time, and responses to experimental conditions---before LLM agents are used as proxies for human decision-makers.

\end{document}